\documentclass{article} 

\usepackage[utf8]{inputenc}

\usepackage{appendix}
\usepackage{url}
\usepackage{multirow}
\usepackage{lipsum}
\usepackage{amsthm}
\usepackage{mathtools}
\usepackage{xfrac}
\usepackage[export]{adjustbox}
\usepackage{longtable}
\usepackage{booktabs}
\usepackage{colortbl}
\usepackage[dvipsnames]{xcolor}
\usepackage{graphicx}
\usepackage{subcaption}
\usepackage{changelog}
\usepackage{chronosys}
\usepackage{verbatim}
\usepackage{placeins}
\usepackage{tikz}
\usetikzlibrary{positioning,shapes,arrows.meta,fit}
\usepackage{fontawesome}
\usepackage{pgfplots}
\usepackage{float}

\usepackage{svg}
\usepackage{hyperref}
\hypersetup{
    colorlinks=true,
    urlcolor=[rgb]{0.188, 0.498, 0.886},
    linkcolor=[rgb]{0.188, 0.498, 0.886},    
    filecolor=magenta,      
    citecolor=[rgb]{0.188, 0.498, 0.886},
}
\usepackage{fancyvrb}
\usepackage{fixltx2e}
\usepackage{bera}
\usepackage{pdfpages}
\usepackage{pgfgantt}
\usepackage{xcolor}

\definecolor{barblue}{RGB}{153,204,254}
\definecolor{groupblue}{RGB}{51,102,254}
\definecolor{linkred}{RGB}{165,0,33}

\usepackage{pgf-umlsd}
\usepackage[draft=true]{minted}

\catcode`\@=11
\def\chron@selectmonth#1{\ifcase#1\or January\or February\or March\or April\or%
 May\or June\or July\or August\or September\or October\or November\or December\fi}

\usepackage{algorithm} 
\usepackage{program} 
\usepackage{algorithmic}
\usepackage{listings}
\usepackage{geometry}
\usepackage{authblk}

\usepackage{caption,setspace}
\DeclareCaptionFormat{listing}{\rule{\dimexpr0.9\columnwidth+17pt\relax}{0.4pt}\par\vskip1pt#1#2#3}
\newcounter{code}



\DeclareMathVersion{sans}
\SetSymbolFont{operators}{sans}{OT1}{cmbr}{m}{n}
\SetSymbolFont{letters}  {sans}{OML}{cmbrm}{m}{it}
\SetSymbolFont{symbols}  {sans}{OMS}{cmbrs}{m}{n}

\lstnewenvironment{sflisting}[1][]
  {\lstset{#1}\mathversion{sans}}{}

\usepackage[normalem]{ulem}

\usetikzlibrary{%
  arrows,%
  shapes.misc,
  shapes.arrows,%
  chains,%
  matrix,%
  positioning,
  scopes,%
  decorations.pathmorphing,
  shadows%
}

\definecolor{grayalias}{HTML}{3F4444}

\definecolor{bluealias}{HTML}{307FE2}

\title{
\textbf{\texttt{ExploitFlow}}, cyber security exploitation routes for Game Theory and AI research in robotics
}

\author[1,2]{
    Víctor Mayoral Vilches
}

\author[3]{Gelei Deng}
\author[3]{Yi Liu}
\author[1]{Martin Pinzger}
\author[4,1]{Stefan Rass}

\affil[1]{\footnotesize \href{https://aau.at/}{Alpen-Adria-Universität Klagenfurt}, Austria}
\affil[2]{\footnotesize \href{https://aliasrobotics.com/}{Alias Robotics}, Spain}
\affil[3]{\footnotesize \href{https://www.ntu.edu.sg/}{Nanyang Technological University}, Singapore}
\affil[4]{\footnotesize \href{https://www.jku.at/}{Johannes Kepler University Linz}, Austria}

\newcommand{\exploitflow}{\texttt{ExploitFlow} }
\newcommand{\exploitflows}{\texttt{ExploitFlow}}

\newcommand{\tool}{\texttt{PentestGPT}}
\newcommand{\framework}{\texttt{Malism}}
\newcommand{\ef}{\texttt{ExploitFlow}}
\newcommand{\pentestperf}{\texttt{PentestPerf}}

\begin{document}
\maketitle


\begin{abstract}


This paper addresses the prevalent lack of tools to facilitate and empower Game Theory and Artificial Intelligence (AI) research in cybersecurity. The primary contribution is the introduction of \exploitflow (EF), an AI and Game Theory-driven modular library designed for cyber security exploitation. EF aims to automate attacks, combining exploits from various sources, and capturing system states post-action to reason about them and understand potential attack trees. The motivation behind EF is to bolster Game Theory and AI research in cybersecurity, with robotics as the initial focus. Results indicate that EF is effective for exploring machine learning in robot cybersecurity. An artificial agent powered by EF, using Reinforcement Learning, outperformed both brute-force and human expert approaches, laying the path for using \exploitflow for further research. Nonetheless, we identified several limitations in EF-driven agents, including a propensity to overfit, the scarcity and production cost of datasets for generalization, and challenges in interpreting networking states across varied security settings. To leverage the strengths of \exploitflow while addressing identified shortcomings, we present \framework{}, our vision for a comprehensive automated penetration testing framework with \exploitflow at its core.

\end{abstract}


\section{Introduction}
\label{sec:motivation}

Robots are often insecure and fully unprotected. The rationale behind this is fourfold: first, defensive security mechanisms for robots are still in their early stages, not covering the complete threat landscape. Second, the inherent complexity of robotic systems makes their protection costly, both technically and economically. Third, robot vendors are currently not taking responsibility in a timely manner, extending the zero-days exposure window (time until mitigation of a zero-day) to several years on average \cite{mayoral2019industrial}. Fourth, contrary to the common‐sense expectations and similar to Ford in the 1920s with cars, most robot manufacturers oppose or difficult robot hardware repairs and software patching. They employ planned obsolescence practices to discourage repairs and evade competition. In addition, it is observed how most manufacturers keep forwarding the cyber security problems to the end-users of these robotic machines, further obstructing the security landscape in robotics. Security is not a product, but a process that needs to be continuously assessed in a periodic manner, as systems evolve and new cyber-threats are discovered. Automation is crucial to tackle this problem. Specially, given the lack of security professionals and how long it takes to train qualified security researchers.

Against the current overwhelming insecurity landscape in robotics, this research paper advocates for offensive security methods for robots. These methods are necessary to understand attackers' behavior, to train defensive mechanisms and ultimately, to help protect existing systems by discovering flaws first. Building upon about a decade of robotics and empowering the use of Artificial Intelligence (AI) and Game Theory to automate attacks, this line of research aims to study how offensive AI-driven cyber security methods apply to robotics and allow to protect such systems in a feasible manner. In particular, this work presents \texttt{ExploitFlow} (\texttt{EF}), a modular library to produce cyber security exploitation routes (\emph{exploit flows}). \texttt{ExploitFlow} aims to combine and compose exploits from different sources and frameworks, capturing the state of the system being tested in a flow after every discrete action which allows learning attack trees that affect a given system.

\exploitflows's main motivation is to facilitate and empower Game Theory and Artificial Intelligence (AI) research in cyber security. A secondary motivation is to put \exploitflow in practice in an area wherein exploitation routes can help significantly secure complex systems via offensive cyber security mechanisms. Robotics is selected as the target area for this research, but other targets might be included in the future. To facilitate adoption, \exploitflows's syntax and architecture is inspired by TensorFlow\cite{abadi2016tensorflow}.


\definecolor{codegreen}{rgb}{0.3,0.6,0.3}
\definecolor{codegray}{rgb}{0.5,0.5,0.5}
\definecolor{codepurple}{rgb}{0.5,0,0.33}
\definecolor{backcolour}{rgb}{0.95,0.95,0.92}

\lstdefinestyle{mystyle}{
    backgroundcolor=\color{backcolour},
    commentstyle=\color{codegreen},
    keywordstyle=\color{blue},
    numberstyle=\tiny\color{codegray},
    stringstyle=\color{codepurple},
    basicstyle=\ttfamily\footnotesize,
    breakatwhitespace=false,
    breaklines=true,
    captionpos=b,
    keepspaces=true,
    numbers=left,
    numbersep=5pt,
    showspaces=false,
    showstringspaces=false,
    showtabs=false,
    tabsize=2
}
\lstset{style=mystyle,caption={\exploitflow usage example.}}
\begin{lstlisting}[language=Python]
import exploitflow as ef

flow = ef.Flow()      # Create a simple exploit flow
a = ef.placeholder()  # Instantiate a simple operation
print(flow.run(a))    # Run the flow and print results
\end{lstlisting}

\noindent A more complex exploitation route example is depicted below which performs reconnaissance locating multiple targets in the local area network and then conducts scans on each one of them to fill up the state:

\lstset{style=mystyle,caption={\exploitflow example doing multi-target reconnaissance. A exploit flow is built that performs reconnaissance locating multiple targets in the local area network and then conducts scans a targeted scan to \texttt{192.168.2.10} to fill up a state object that can then be used for reasoning.}}
\begin{lstlisting}[language=Python]
import exploitflow as ef

flow = ef.Flow()
init = ef.Init()
recon = ef.Targets()  # Build MSF reconnaissance exploit
versions = ef.Versions(ports=ef.state.TARGET_PORTS_COMPLETE)

# initialize state and pass it over a recon action
# resulting flow should deliver a state annotated
# with the results from the reconnaissance step
state = flow.run(init * recon * versions, target="192.168.2.10")
\end{lstlisting}

To simplify exploitation, \exploitflow represents each action in an exploitation route with the superclass Exploit, which includes reconnaissance and control actions\footnote{We are well aware that, strictly speaking, reconnaissance scripts don't meet the formal definition of an exploit. However, we still insist on grouping all actions under the same common class (Exploit) to simplify the production of exploitation flows (routes).}. Exploits are grouped into six major categories, inspired by the security kill chain\cite{lockheed2011}. \exploitflow is a modular, extensible library that accepts connectors for other exploitation frameworks and/or individual exploits, and is composable. \exploitflow is not an exploitation framework, but a tool to produce cybersecurity exploitation routes, which empowers research in Game Theory and Artificial Intelligence (AI) in cybersecurity.\\

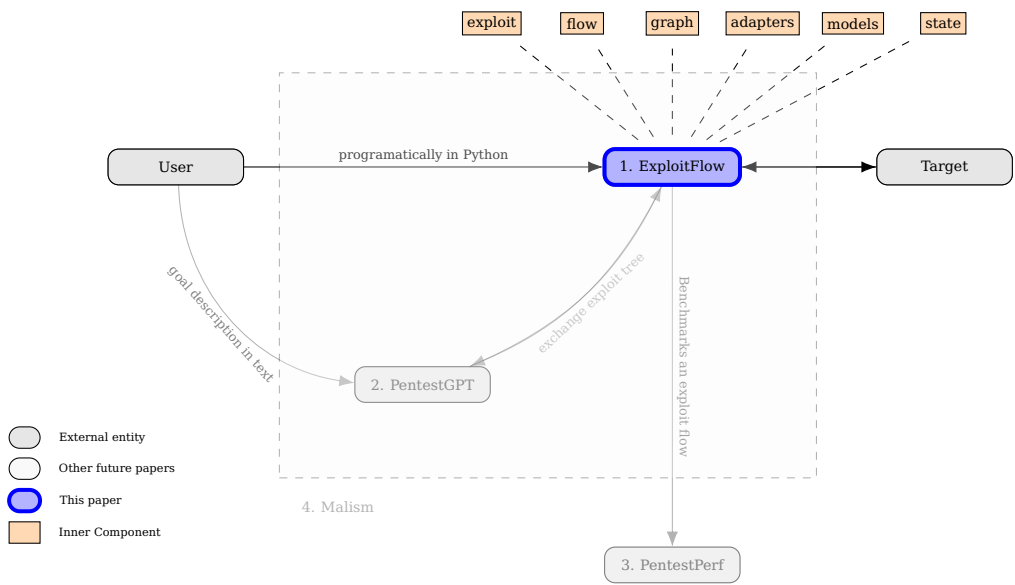
\begin{figure}[b!]
    \centering 
    \begin{tikzpicture}[node distance=2cm, auto, scale=0.6, transform shape]
    \tikzstyle{box} = [rectangle, draw, fill=gray!20, text centered, rounded corners, minimum height=2.5em, minimum width=3cm]
    \tikzstyle{emphasize} = [box, fill=blue!30, draw=blue, line width=1.5pt]
    \tikzstyle{line} = [draw, -{Latex[scale=1.2]}]
    \tikzstyle{deemphasize} = [opacity=0.3]
    \tikzstyle{innerComp} = [rectangle, draw, fill=orange!30, text centered, minimum height=1.5em, minimum width=3em]

    \node[box] (U) {User};
    \node[emphasize] (EF) at (11cm,0) {1. ExploitFlow}; 
    \node[box, deemphasize, below left=4cm and 2.5cm of EF] (PentestGPT) {2. PentestGPT};
    \node[box, deemphasize, below=8cm of EF] (PentestPerf) {3. PentestPerf};
    \node[box, right=3cm of EF] (T) {Target};

    \node[innerComp, above=2.5cm of EF, xshift=-4cm] (exploit) {exploit};
    \node[innerComp, right of=exploit] (flow) {flow};
    \node[innerComp, right of=flow] (graph) {graph};
    \node[innerComp, right of=graph] (adapters) {adapters};
    \node[innerComp, right of=adapters] (models) {models};
    \node[innerComp, right of=models] (state) {state};
    
    \draw[dashed, shorten <=5pt, shorten >=5pt] (EF) -- (exploit);
    \draw[dashed, shorten <=5pt, shorten >=5pt] (EF) -- (flow);
    \draw[dashed, shorten <=5pt, shorten >=5pt] (EF) -- (graph);
    \draw[dashed, shorten <=5pt, shorten >=5pt] (EF) -- (adapters);
    \draw[dashed, shorten <=5pt, shorten >=5pt] (EF) -- (models);
    \draw[dashed, shorten <=5pt, shorten >=5pt] (EF) -- (state);

    \path[line] (U) -- node[font=\small, midway, sloped, above] {programatically in Python} (EF); 
    \path[line, deemphasize] (U) edge[bend right=40] node[font=\small, sloped, below, opacity=0.5] {goal description in text} (PentestGPT);
    \path[line, deemphasize] (PentestGPT) edge[bend right=20] node[font=\small, sloped, below] {exchange exploit tree} (EF);
    \path[line, deemphasize] (EF) edge[bend left=20] node {} (PentestGPT);
    \path[line, deemphasize] (EF) -- node[font=\small, sloped, above, opacity=0.5] {Benchmarks an exploit flow} (PentestPerf);
    \path[line] (EF) -- (T);
    \path[line] (T) -- (EF);

    \node[rectangle, fill=gray!5, draw, dashed, deemphasize, fit=(EF) (PentestGPT), inner sep=1cm] (group) {};
    \node[anchor=north west, deemphasize, inner sep=0.5cm] at (group.south west) {4. Malism};

    \node[box] (U) {User};
    \node[emphasize] (EF) at (11cm,0) {1. ExploitFlow}; 
    \node[box, deemphasize, below left=4cm and 2.5cm of EF] (PentestGPT) {2. PentestGPT};
    \node[box, deemphasize, below=8cm of EF] (PentestPerf) {3. PentestPerf};
    \node[box, right=3cm of EF] (T) {Target};

    \begin{scope}[node distance=0.3cm, font=\footnotesize]
        \coordinate (legendOrigin) at (-4,-6); 
        \node[box, right=0.3cm of legendOrigin, minimum height=1.5em, minimum width=1.5em, text width=1.5em] (legendBox) {}; 
        \node[right=0.3cm of legendBox] {External entity};

        \node[box, below=0.2cm of legendBox, fill=gray!5, minimum height=1.5em, minimum width=1.5em, text width=1.5em] (otherpapers) {}; 
        \node[right=0.3cm of otherpapers] {Other future papers};

        \node[emphasize, below=0.2cm of otherpapers, minimum height=1.5em, minimum width=1.5em, text width=1.5em] (legendEmphasize) {};
        \node[right=0.3cm of legendEmphasize] {This paper};

        \node[innerComp, below=0.2cm of legendEmphasize, minimum height=1.5em, minimum width=1.5em, text width=1.5em] (legendInner) {};
        \node[right=0.3cm of legendInner] {Inner Component};
    \end{scope}
\end{tikzpicture}
    \caption{Architecture of our framework to develop a fully automated penetration testing tools, \framework{}. Figure depicts the various interaction flows that an arbitrary \texttt{User} could follow using \framework{} to conduct penetration testing (pentest) a given \texttt{Target}. \textbf{1. (this paper)} Corresponds with \ef{}, a modular library to produce security exploitation routes (\emph{exploit flows}) that captures the state of the system being tested in a flow after every discrete action. \textbf{2.} Corresponds with \tool{}, a testing tool that leverages the power of LLMs to produce testing guidance (heuristics) for every given discrete state. \textbf{3.} \pentestperf is a comprehensive penetration testing benchmark to evaluate the performances of penetration testers and automated tools across a wide array of testing targets. \textbf{4.} captures \framework{}, our framework to develop fully automated penetration testing tools which we name \emph{cybersecurity cognitive engines}.
    }
    \label{fig:my_figure_label} 
\end{figure}

\noindent With this understanding, \exploitflow aims to contribute to unlocking the potential of modern machine learning approaches and developing a fully automated penetration testing framework that helps produce cybersecurity cognitive engines. Our overall architecture is depicted in Figure~\ref{fig:my_figure_label}, which shows our work thus far and our planned contributions for the near future. Our proposed framework, \framework{}, is designed to enable a user without in-depth security domain knowledge to create their own cybersecurity cognitive engine that helps conduct penetration testing over an extensive range of targets. This framework comprises three primary components:
\begin{enumerate}

\item \ef{} (\textbf{this paper}): A modular library to produce cyber security exploitation routes (\emph{exploit flows}). \ef{} aims to combine and compose exploits from different sources and frameworks, capturing the state of the system being tested in a flow after every discrete action which allows learning attack trees that affect a given system. \ef{}'s main motivation is to facilitate and empower Game Theory and Artificial Intelligence (AI) research in cyber security. It provides a unique representation of the exploitation process that encodes every facet within it. Its representation can be effectively integrated with various penetration testing tools and scripts, such as Metasploit~\cite{moore2003metasploit} to perform end-to-end penetration testing. Such a representation can be further visualized to guide the human experts for the reproduction of the testing process. 

\item \tool{}: An automated penetration testing system that leverages the power of LLMs to produce testing guidance and intuition at every given discrete state. It functions as the core component of the \framework{} framework, guiding the LLMs to efficiently utilize their domain knowledge in real-world testing scenarios.

\item \pentestperf: A comprehensive penetration testing benchmark developed to evaluate the performance of penetration testers and automated tools across a wide array of testing targets. It offers a fair and robust platform for performance comparison.
\end{enumerate}

\noindent The harmonious integration of these three components forms an automated, self-evolving penetration testing framework capable of executing penetration tests over various targets, \framework{}. This framework to develop fully automated penetration testing tools, which we name \emph{cybersecurity cognitive engines}, aims to revolutionize the field of penetration testing by significantly reducing the need for domain expertise and enabling more comprehensive and reliable testing.\\

Contributions of this article are two-fold: first, we present for the first time the implementation and use of \exploitflow as a tool to build security exploitation routes. Second, we present the results obtained while creating an artificial agent powered by \exploitflow, that is able to locate exploits affecting the various robotic systems presented to the agent, while minimizing the amount of network traffic and attacks generated (exploits attempted). Q-Learning is used as the learning model and compared against two baselines: a) brute force and b) human expert. The source code including both the \exploitflow implementation as well as the machine learning experiments are available at \faGithub~  \url{https://github.com/vmayoral/ExploitFlow}.\\

Results presented below in section \ref{sec:evaluation} hint that \exploitflow is usable and useful for exploring ML use in the context of robot cybersecurity. Three actors are evaluated: the brute-force actor obtains a {\color{red!30!white}\texttt{-2680}} cumulative reward (see \ref{sec:learningmodel} for an explanation of the reward structure), the human expert {\color{blue!30!white}\textbf{\texttt{8}}} and the autonomous agent using value-based Reinforcement Learning (RL) {\color{green!30!white}\textbf{\texttt{100}}}. Attacks graphs shown hint that while the human expert iterates over reconnaissance exploits to introspect the networking scenario, the agent directly learns to submit the exploit that triggers a favourable reward and then iterates over an Idle action to maximize resulting reward (obtaining maximum possible, in fact). This overfitting behavior is a shortcoming to tackle in future work, since it impedes generalization across new testing environments. Overall, \exploitflow is demonstrated by creating an artificial agent powered by \exploitflow that is able to locate exploits affecting the target robotic systems presented to the agent, using value-based Reinforcement Learning (Q-Learning ) and while minimizing the amount of network traffic and attacks generated (exploits attempted). \exploitflow also demonstrates how the state of the system being tested is captured after every discrete action, which allows reasoning, keeping track of action/state pairs, and producing attacks trees affecting a given system. Ultimately, we present in Figure \ref{fig:my_figure_label} the architecture of our framework to develop a fully automated penetration testing tools, \framework{}, which we name cybersecurity cognitive engines, and which uses \exploitflow at its core.

\section{Methodology}
\label{sec:methodology}


This section describes the research approach, including the data used for the learning task and the learning model. Subsection~\ref{cornerstones} will first set the context, the current state of the art and will highlight the lack of available datasets for robot cybersecurity study. Section~\ref{datagen} will then discuss our approach for data generation through CTF-like OS-virtualized environments to generate realistic robotics networking data and what kind of preprocessing and other manpulations were applied to it for the learning task. Finally, Section~\ref{sec:learningmodel} will present an overview of our learning approach and the model used.

\newpage

\subsection{Biographical cornerstones in machine learning applied methods to robot cybersecurity}
\label{cornerstones}

The following presents a summary of some of the most relevant and recent research articles related to offensive cybersecurity approaches that leverage machine learning:

\vspace{-4cm}

\startchronology[startyear=2005,stopyear=2025,height=1.0ex]

\chronoevent[textwidth=7cm,markdepth=140pt]{08/2022}{\small Hierarchical reinforcement learning for efficient and effective automated penetration testing of large networks \cite{ghanem2022hierarchical}}

\chronoevent[textwidth=8cm,markdepth=-150pt]{2020}{\small \bf Modeling penetration testing with reinforcement learning using capture-the-flag challenges: trade-offs between model-free learning and a priori knowledge \cite{zennaro2020modeling}}

\chronoevent[textwidth=3cm,markdepth=10pt]{2005}{\small MulVAL: A Logic-based Network Security Analyzer \cite{ou2005mulval}}

\chronoevent[textwidth=5cm,markdepth=-20pt]{2013}{\small PEGASUS: A policy search method for large MDPs and POMDPs \cite{ng2013pegasus}}

\chronoevent[textwidth=3cm,markdepth=25pt]{2016}{\small Intelligent, automated red team emulation \cite{applebaum2016intelligent}}

\chronoevent[textwidth=4cm,markdepth=85pt]{09/2019}{\small NIG-AP: a new method for automated penetration testing \cite{zhou2019nig}}

\chronoevent[textwidth=6cm,markdepth=-95pt]{12/2019}{\small Reinforcement learning for efficient network penetration testing \cite{ghanem2019reinforcement}}

\chronoevent[textwidth=4cm,markdepth=-30pt]{2020}{\small \bf Automated penetration testing using deep reinforcement learning \cite{hu2020automated}}

\chronoevent[textwidth=3cm,markdepth=20pt]{2021}{\small Automating post-exploitation with deep reinforcement learning \cite{maeda2021automating}}

\chronoevent[textwidth=8cm,markdepth=200pt]{2023}{\small GAIL-PT: An intelligent penetration testing framework with generative adversarial imitation learning \cite{chen2023gail}}

\stopchronology

System complexity is the enemy of security. It is very difficult to assure no vulnerabilities in a system that an attacker could exploit. In robotics, the reality is even more concerning, as robots are complex systems, with wide attack surfaces and wherein there is no culture of security yet. As studied in previous work~\cite{2018arXiv180606681A, vilches2019introducing}, except few vendors and selected actions, no real concern has been shown for security in robotics. Offensive security methods (such as pentesting or red teaming) play an important part in the security lifecycle. In these engagements, as introduced by Applebaum~\cite{applebaum2017analysis}, security teams try to break into an organization’s assets, identifying vulnerabilities along the way. Red teams take this concept even further, trying to fully emulate what real adversaries do: instead of just compromising the network and identifying vulnerabilities, they have a larger goal that requires significant post-compromise work.

Various authors, including those summarized above took note of this, and started leveraging offensive mechanisms to tackle the cybersecurity problem. In particular, \cite{ghanem2018reinforcement, niculaelearning2018, schwartz2019autonomous} among others started using machine learning to automate the penetration testing task, presenting different levels of automation. Most of the authors use rather simple  neural network architectures, many leveraging Deep Q-Learning (DQN) and mostly involving few hidden layers, besides the input and output ones. This hints about the fact that the tasks learned are rather simple, and a simplification of the complete penetration testing activities. Something coherent across most of these studies is the lack of consistency concerning the representation of the state space for cybersecurity research purposes in ML. Connected to it, we found that most cited authors above use inconsistent datasets.  Though various popular \emph{security-oriented networking datasets} exist (see \faGithub~  \url{https://github.com/shramos/Awesome-Cybersecurity-Datasets}) most of the authors reviewed seem to discard these and instead choose to generate their own data via either abstractions (simplified models) or simulated/emulated environments. Further research hinted that rationale behind this is two-fold:

\begin{enumerate}
    \item {Capturing networking data in a scalable manner is non-trivial (and an open problem)}
    \item {Cybersecurity environments are highly non-structured and interactive}
\end{enumerate}

In this work, we fail to locate a dataset that could serve the research goals. Deploying offensive cybersecurity approaches in the complex robotics field presents quite a challenge, given its novelty. Also, because the robotics realm is prohibitive: cost, repeatability, and expertise all make it difficult to consistently produce offensive security datasets. Following from this, and similar to some of the cited work above. To proceed with this research, we attempted addressing the two identified hurdles cited above by collecting synthetic data directly from a realistic OS-virtualized scenario with multiple robotic targets. The following sections discuss our approach.

\subsection{Synthetic data generation through CTF-like OS-virtualized environments}
\label{datagen}

\subsubsection{Data generation}

Obtaining a proper dataset is crucial and specially hard in certain areas, such as the unstructured robotics or cybersecurity. The intersection of both presents an even more difficult challenge. In particular for security, automating offensive cyber security practices constitutes a non‐trivial problem because of the range and complexity of actions that a human expert may attempt, which hampers putting together representative datasets. The authors of \cite{zennaro2020modeling} acknowledge this and focus their attention on simplified penetration testing problems expressed in the form of Capture The Flag (CTF) hacking challenges. They tackle the \emph{dataset production} problem by leveraging existing CTF environments, which they modify to adapt to modern RL \emph{gyms} which produce a reward after an agent performs an action. This way, authors are able to \emph{produce} relevant input data (which serves as datasets) for the learning offensive cyber security practices in selected target scenarios. The research herein will reproduce the approach followed at \cite{zennaro2020modeling} with relevant  extensions meant for modeling robots. The reader should note that while pure networking (for security purposes) modeling can easily be done with virtualization (e.g. OS-virtualization such as Docker, or VMs which are commonly used), robotic environments require a significantly more complex abstraction set, which elevates the complexity of building such CTF environments. To address this problem, the present work builds on top of some of \cite{mayoral2020alurity}, which provides a toolbox for robot cybersecurity concerning. Altogether, we are able to build OS-virtualized CTF environments with the target robots that allow performing RL on them. 

Data is thereby generated as synthetic and directly from the simulated robotic environments, each of which can fit a different challenge, or CTF game. The data corresponds to the simulated networking interactions between peers. The following describes how such networking dataflow is abstracted away for learning purposes.

\subsubsection{Data pre-processing}

Learning from computer networking data requires relevant data pre-processing, as the amount of information is overwhelming. The following presents our initial assumptions while considering networking data pre-processing. 

Let's assume the following:
\begin{itemize}
    \item At each time-step, the network is captured by a \texttt{State} class. On its simplest form, each \texttt{State} is represented by a dictionary, each key corresponding with an IP address and holding a sub-state class which captures (for each IP) a) exploits launched against the network and b) port status (open/closed, versions, cpe, etc.).
    \item One-hot encoding is used as the safest approach to interface with the learning model. Others considered included label encoding and binary encoding. While the latter produced shorter representations, binary encoding might not be suitable for all machine learning algorithms. In binary encoding, although categories are converted into numbers, there is an arbitrary ordering introduced in the categories, which might mislead some algorithms into thinking that there is a relationship between different categories. One-hot encoding instead  doesn't have this issue as it uses binary vectors to represent categories, at the cost of increasing the dimensionality of the dataset.        
\end{itemize}

Given these assumptions, we considered initially the following:

\begin{itemize}
    \item $y = 255$, number of IPs considered (a complete minor subnet)
    \item $n = 424$, number of ports monitored, each represented by $l = 1$ bit to capture its state (open/close)
    \item $m = 12$, number of exploits considered, each represented by $b = 1$ bits (launched/not-launched), discarding capturing the success/failure of each exploit to model better the uncertainty often encountered in cybersecurity.
    \item $s = 0$ bits capturing the system information for each IP (OS version, and other meta-data obtained while doing reconnaissance)
\end{itemize}

These assumptions, which already introduce a significant simplification, while one-hot encoded already lead to the following number of elements while encoding the \texttt{State}:

\begin{equation*}
    \begin{split}
y \cdot [n \cdot (n + l) + \cdot (m + b) + s] = \\
255 \cdot [424 \cdot (424 + 1) + 12 \cdot (12 + 1) + 0] = \\
256 \cdot 180,356 = 46,171,136
    \end{split}
\end{equation*}

This leads to a setup which is non-feasible computationally. These initial assumptions had to change. We had to significantly increase abstractions (and reduce complexity in terms of network-scope) to make computations tractable in modern general-purpose computers. In particular we used $y = 7$, $n = 9$, $l = 1$, $m = 12$, $b = 1$ and $s = 0$ leading to:

\begin{equation*}
    \begin{split}
y \cdot [n \cdot (n + l) + \cdot (m + b) + s] = \\
7 \cdot [9 \cdot (9 + 1) + 12 \cdot (12 + 1) + 0] = \\
7 \cdot 246 = 1,722
    \end{split}
\end{equation*}

Preserving the modularity principles of \exploitflows, all the above is implemented by a \texttt{State} class that can easily be extended, overwritten and/or modified per each experiment independently. The listing \ref{lst:state} shows an example implementing a variant of \texttt{State} called \texttt{State\_v2}:


\subsection{Learning model}  
\label{sec:learningmodel}

As described above, most of the prior art reviewed use rather simple neural network architectures for cybersecurity automation and machine learning, many leveraging Deep Q-Learning (DQN) and mostly involving few hidden layers, besides the input and output ones. In a way, this could be understood as an indicator that the field is still mostly immature (from an ML-perspective) and that simple problems are being explored, which don't demand complex models. In light of this and particularly following the trend established by \cite{zennaro2020modeling} which focus on scalability, rather than complexity of the problem to solve using a more \emph{explainable} model, we adopt a similar approach and leverage reinforcement learning with simple table-based Q-Learning as a model.\\
\newline
Besides Q-Learning, the following hyperparameters are used consistently across our machine learning efforts: learning rate (alpha, $\alpha$) = 0.1, discount factor (gamma, $\gamma$) = 0.9, exploration factor (epsilon, $\epsilon$) = 0.1.\\
\newline
Finally, the following reward scheme was applied:
\begin{itemize}
    \item Non-exploits, like idle (no action) and related were assigned a reward of $0$, helping the agent to learn that idle might be appropriate given certain networking conditions.
    \item reconnaissance (both fingerprinting and footprinting) actions were assigned a reward of $-10$ by default, and an extra $-1$ for each IP target impacted while incurring in footprinting. This way, we penalize the additional networking traffic generated, which can help Intrussion Detection Systems (IDS) detect malicious activity.
    \item Formal offensive exploits were assigned a reward of $100$ when successful, and $-100$ if failed.    
\end{itemize}

\section{Evaluation}
\label{sec:evaluation}

Evaluation was performed in a simulated scenario involving the following robotic systems presented in Figure \ref{fig:scenario}. Target objective of the learning effort is to compromise the Universal Robots UR3 collaborative manipulator using \href{https://github.com/aliasrobotics/RVD/labels/robot%20component:%20Universal%20Robots%20Controller?page=4&q=is%3Aopen+label%3A%22robot+component%3A+Universal+Robots+Controller%22}{well known security vulnerabilities} affecting this robotic system. In particular, an exploit for compromising the robot using the \href{https://github.com/aliasrobotics/RVD/issues/672}{RVD\#672} (hard-coded public credentials for controller) vulnerability will be used. Three situations were considered and described below:

\begin{figure}[h!]
    \includegraphics[width=0.8\textwidth]{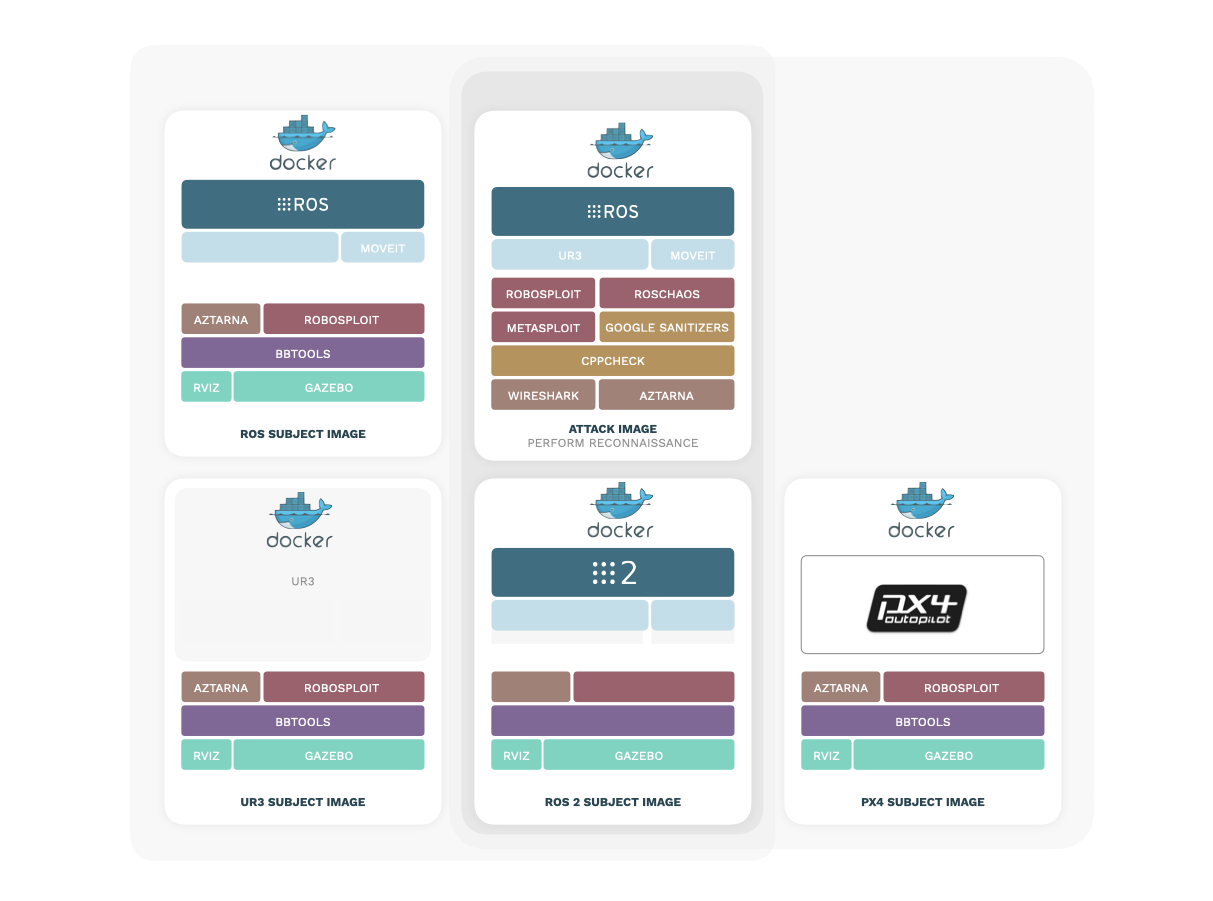}
    \centering
    \caption{CTF-like OS-virtualized robotic environments involving 4 robotic targets: a robotic brain powered by ROS 2, another powered by ROS, a UR3 cobot manipulator and a PX4 drone. Source code to reproduce this environment is available at \faGithub~  \url{https://github.com/vmayoral/ExploitFlow/blob/main/.devcontainer/docker-compose.yml}}
    \label{fig:scenario}
\end{figure}


\subsection{Human-expert penetration tester}

A human-expert was considered by using \exploitflow programatically, and manually programming an exploitation route for the best-case scenario. Exploitation route is achieved using the code snippet at listing \ref{lst:human} (complete experiment available at \faGithub~  \url{https://github.com/vmayoral/ExploitFlow/blob/main/examples/9_exploitation_ur_human_expert.py}).

\begin{figure}[h!]
    \includegraphics[width=0.8\textwidth]{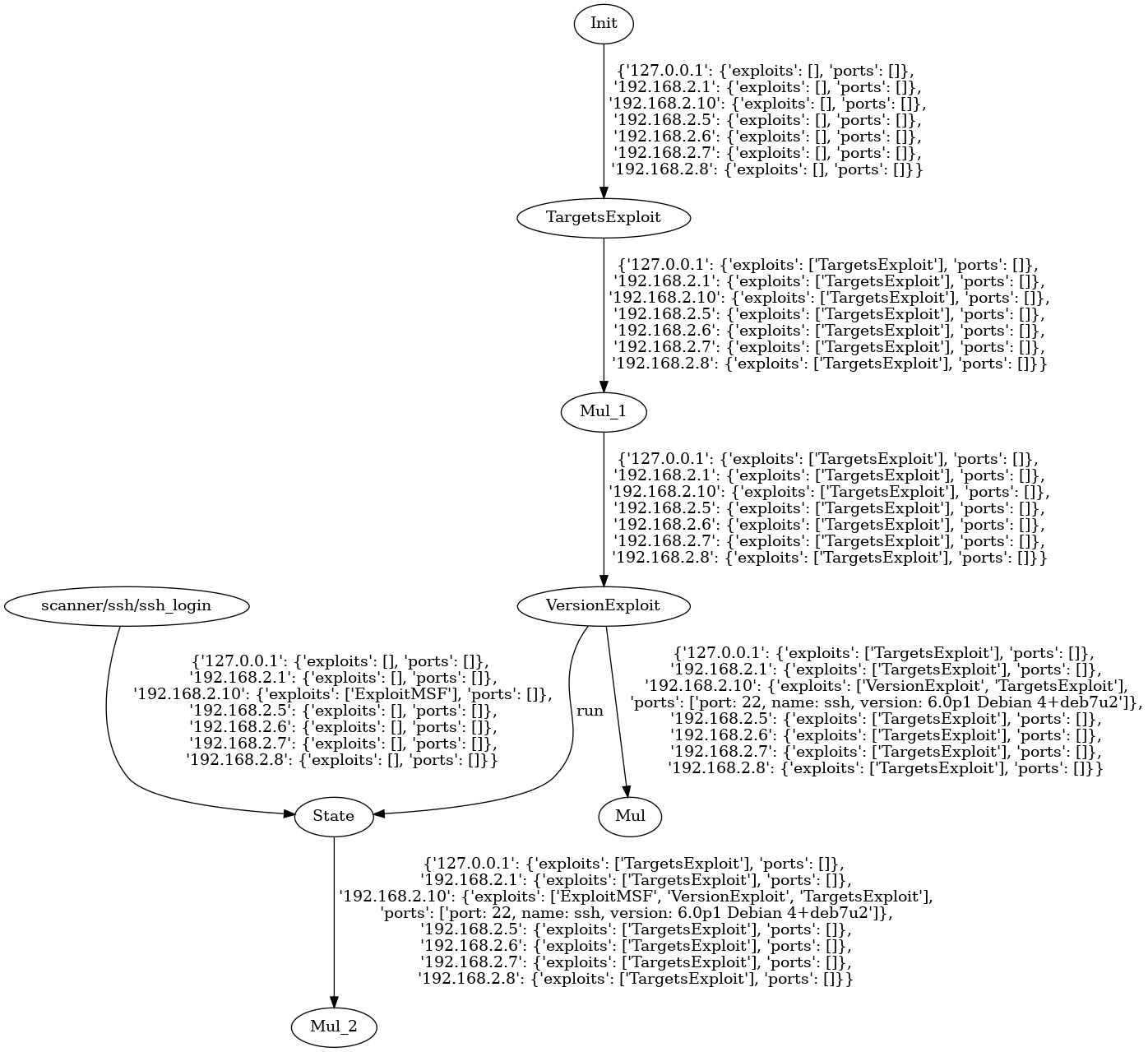}
    \centering
    \caption{Depiction of the attack graph produced by a human expert while executing the manually-written best-case exploit flow.}
    \label{fig:humanexpert}
\end{figure}

The execution of this exploitation flow leads to the attack graph depicted in Figure \ref{fig:humanexpert} and led to a cumulative reward of $8$.

\subsection{Autonomous agent powered by Q-Learning}

An artificial agent powered by \exploitflow is able to locate exploits affecting the target robotic system, while minimizing the amount of network traffic and attacks generated (exploits attempted) using its table-based Q-Learning model. Exploitation route is learned using the code snippet at listing \ref{lst:agent} (complete experiment available at \faGithub~  \url{https://github.com/vmayoral/ExploitFlow/blob/main/examples/11_exploitation_ur_qlearning_instances.py}).

\begin{figure}[h!]
    \includegraphics[width=0.8\textwidth]{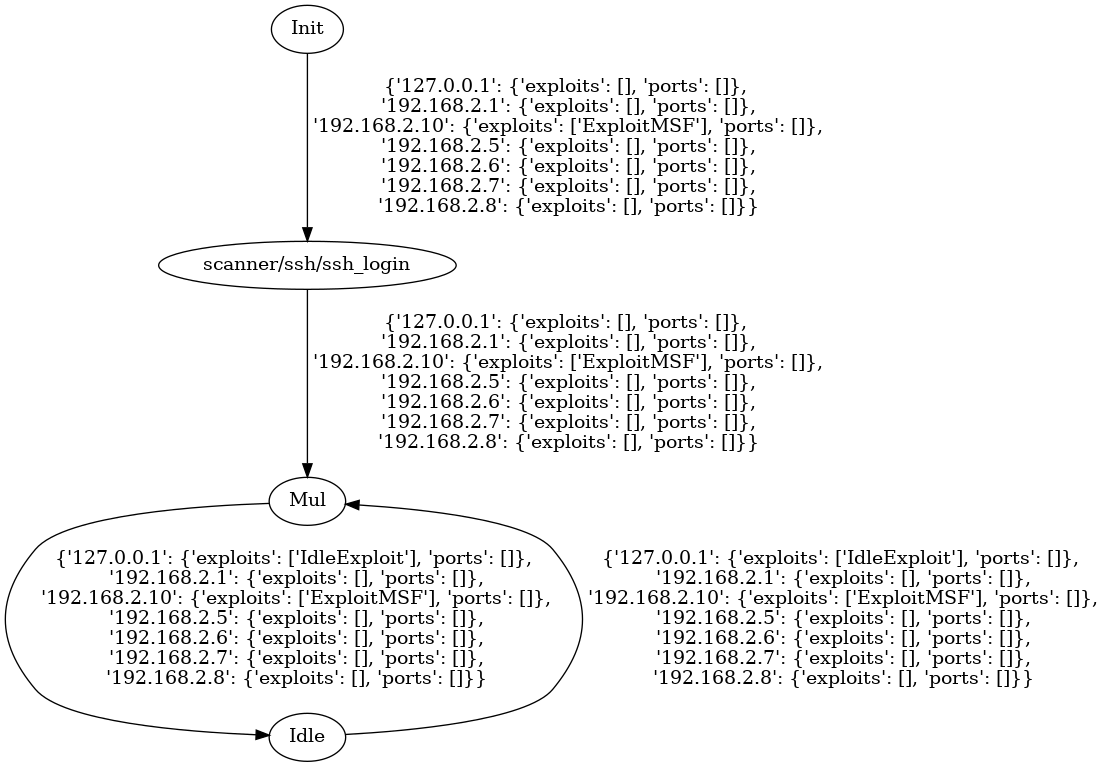}
    \centering
    \caption{Depiction of the attack graph produced by an autonomous agent powered by Q-Learning algorithm while executing the exploit flow that is derived from its learned value-function (table).}
    \label{fig:agent}
\end{figure}

The execution of this exploitation flow leads to the attack graph depicted in Figure \ref{fig:agent} and led to a cumulative reward of $100$.

Implementation was performed in Python, as an extension of \exploitflow (which is coded in Python as well) and in a separate file with standardized interfaces for use. Models abstracted using these interfaces can easily be swapped and tested. Our table-based Q-Learning model presents the form depicted in listing \ref{lst:qlearning}. An extension of this implementation to use other models that leverage artificial neural networks (e.g. DQN) is judged trivial if/when leveraging popular machine learning frameworks such as \emph{TensorFlow}.

\subsection{Brute-force}

A brute force baseline (which often the case relates to what many novel pentesters do) is generated by launching all permutations of the selected available exploits against the scenario\footnote{In fact, a reduced "relevant" set was selected to make it computationally more tractable}. Exploitation route is created using the code snippet available at listing \ref{lst:brute} (complete experiment available at \faGithub~  \url{https://github.com/vmayoral/ExploitFlow/blob/main/examples/13_exploitation_ur_bruteforce.py}). The execution of this exploitation flow leads to a big attack graph (not shown here) and led to a cumulative reward of $-2680$.


\section{Conclusions and future work}
\label{sec:conclusions}

In this article, we made notable contributions to the realm of robot cybersecurity by producing \exploitflow, a modular library to produce cyber security exploitation routes (\emph{exploit flows}) that allows combining and compose exploits from different sources and frameworks, and captures the state of the system being tested in a flow after every discrete action. This allows for further automated reasoning (by means of ML and Game Theory), as well as learning attack trees that affect a given system. Firstly, we introduced the novel implementation of \exploitflow, an innovative tool devised for crafting security exploitation routes. Furthermore, we successfully demonstrated \exploitflow by developing an artificial agent with it, that capably identifies exploits in various robotic systems, significantly minimizing both network traffic and the number of exploits attempted.

Using Q-Learning, we compared the efficiency of our model against two baselines: brute force and human expert. Interested readers can delve into the source code and machine learning experiments available at \faGithub~  \url{https://github.com/vmayoral/ExploitFlow}. \exploitflow's ability to capture the system state post every discrete action is paramount. It aids in logical reasoning, tracing action/state combinations, and generating attack trees for a specific system. To encapsulate our endeavors, Figure \ref{fig:my_figure_label} outlines the architecture of \framework{}, our automated penetration testing tool—termed as a cybersecurity cognitive engine—with \exploitflow as its foundational element.

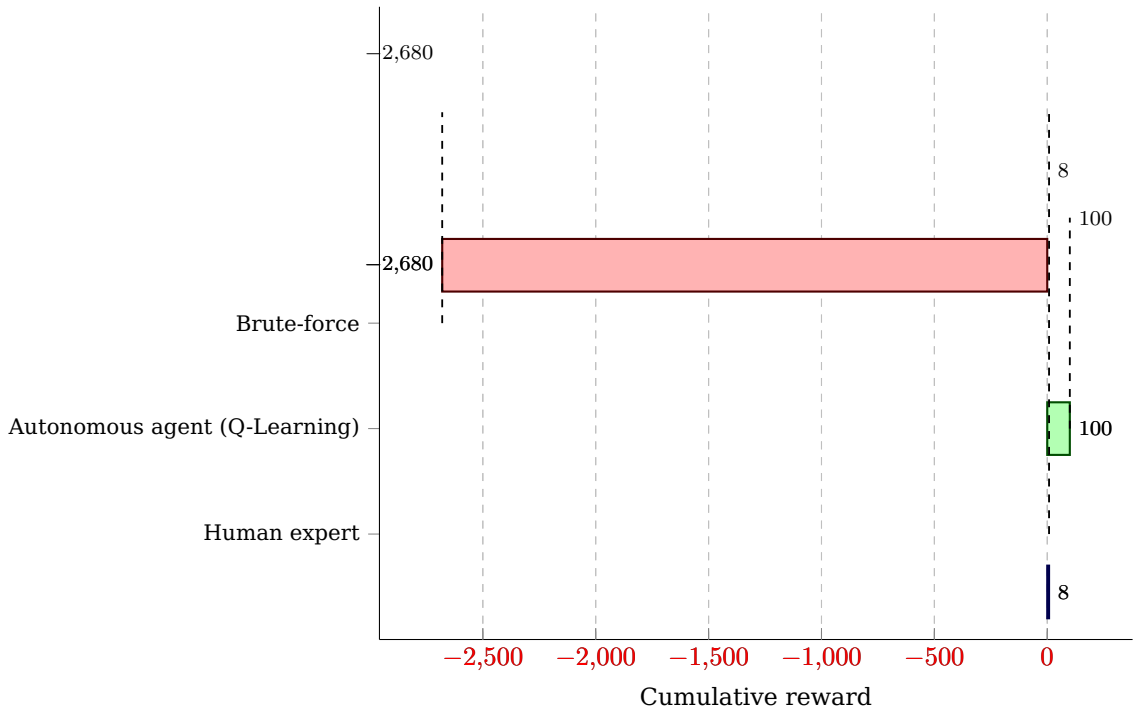
\begin{figure}[h!]
\centering
\begin{tikzpicture}
\begin{axis}[
    xbar,
    xlabel={Cumulative reward},
    xlabel style={align=center},
    xmajorgrids=true, 
    grid style=dashed,
    ytick={1,2,3},
    yticklabels={Human expert, Autonomous agent (Q-Learning), Brute-force},
    nodes near coords,
    nodes near coords align={horizontal},
    nodes near coords style={font=\footnotesize},
    y tick label style={align=center, font=\small},
    cycle list={
        {fill=blue!30!white,draw=blue!30!black, line width=0.8pt}, 
        {fill=green!30!white,draw=green!30!black, line width=0.8pt}, 
        {fill=red!30!white,draw=red!30!black, line width=0.8pt}, 
    },
    height=10cm,
    bar width=0.7cm,
    enlarge y limits={abs=1},
    axis x line*=bottom, 
    axis y line*=left,
    extra x ticks={-2500,-2000,-1500,-1000,-500,0},  
    extra x tick style={tick label style={color=red}},
]
\addplot coordinates {(8,1)};
\addplot coordinates {(100,2)};
\addplot coordinates {(-2680,3)};

\addplot[dashed,sharp plot, line width=0.7pt] coordinates {(8,1) (8,5)} node [right] {};
\addplot[dashed,sharp plot, line width=0.7pt] coordinates {(100,2) (100,4)} node [right] {};
\addplot[dashed,sharp plot, line width=0.7pt] coordinates {(-2680,3) (-2680,5)} node [right] {};
\end{axis}
\end{tikzpicture}
\caption{Cumulative reward obtained by three different actors (brute-force, Q-Learning, human expert) while attempting to compromise the target objective in the robot cybersecurity scenario of figure \ref{fig:scenario}}
\label{fig:cum}
\end{figure}


\noindent Results of the three actors evaluated during this research are summarized in Figure \ref{fig:cum}, depicting the cumulative reward of the exploit flow generated for each actor, respectively (see \ref{sec:learningmodel} for an explanation of the reward structure). The brute-force actor obtains a {\color{red!30!white}\texttt{-2680}} cum. reward, the human expert {\color{blue!30!white}\textbf{\texttt{8}}} and the autonomous agent using value-based Reinforcement Learning (RL) {\color{green!30!white}\textbf{\texttt{100}}}. While analyzing figures \ref{fig:humanexpert} and \ref{fig:agent} it becomes apparent that while the human expert iterates over reconnaissance exploits to introspect the networking scenario, the agent directly learns to submit the exploit that triggers a favourable reward and then iterates over an Idle action to maximize the resulting reward (obtaining maximum possible, in fact). In light of results, we can conclude that the initial research objectives have been fulfilled. \exploitflow implementation was shown functional and demonstrated by creating an artificial agent powered by \exploitflow that is able to locate exploits affecting the target robotic systems presented to the agent, using value-based Reinforcement Learning (Q-Learning ) and while minimizing the amount of network traffic and attacks generated (exploits attempted).\\
\newline
Though results meet the initial goals, we must be critical towards identified shortcomings. The following hint a few directions werein future work could be allocated for further improvements:
\begin{itemize}
    \item First, current learned functions (value-based) overfit to the particular landscape of exploits available and would perform poorly the moment the networking configuration changes (e.g. IP addresses change). To overcome this,  training should be performed without the use of heuristics or simplifications and using \emph{changing} networking setups. Though this can be automated (e.g. creating varios CTF-like scenarios), it was beyond the scope of the current effort. A side-effect of this would be the explosion of the state-space, which would probably require to leverage other function representation models, such as artificial neural networks (as opposed to the current table).
    \item Second, as part of this \emph{overfitting} behavior mentioned, we acknowledge that it is unrealistic for the current autonomous agent to perform properly in the wild, as offensive cybersecurity always requires some level of (continued) reconnaissance. The current agent has not learned to do so. Reward shaping and various scenarios could help addressing this limitation.
    \item Third, the current one-hot encoding is not scalable (computationally very expensive) and alternatives must be explored if the networking \texttt{State} is to be better captured.
    \item Fourth, a better representation of the networking state at each timestamp is required to further scale learning. Ideas in this direction include incorporating machine-related metadata (OS, library versions extracted from finger- and foot-printing).
    \item Fifth and much relevance, there is a huge scalability problem (very costly engineering-wise) while generating datasets that helps autonomous agents learn which action to take next. Creating usable (by ML agents) exploits and scenarios takes a lot of expertise and time. Building appropriate scenarios requires qualified roboticists that understand the dynamics of robotic systems. Adapting exploits for use (or worse, creating new ones) requires cybersecurity experts to craft the right interfaces for automated learning use. To overcome this limitation, modern and popular Large Language Models (LLMs) can be utilized to produce both scenarios as well as exploits in the form/format required, significantly simplifying the effort to produce usable datasets and actions. Listing \ref{lst:llm} shows a hint about how LLMs could be leveraged to tackle this shortcoming. We refer interested readers to submodule 2. \tool{} of Figure \ref{fig:my_figure_label} for more details on this line of work.
    \item Sixth, and somewhat connected to the previous one, the heterogeneity of the security networking environments seems to require heuristics to interpret the networking \texttt{State} and take initial actions while tackling new security challenges in unseen environments. This is somewhat connected to the fact that it is unrealistic to capture all know-how concerning offensive attempts, as most is non-disclosed or in a format that can easily be digested and converted into datasets.  LLMs could also help on this regard, which is a topic worth researching.     
\end{itemize}

Ongoing efforts to tackle these shortcomings are being developed by the authors. Some available in-the-open and publicly (e.g. at \faGithub~ \url{https://github.com/GreyDGL/PentestGPT}). We also refer interested readers to Figure~\ref{fig:my_figure_label} to get a hint of the overall architecture of \framework{}, our framework to produce cybersecurity cognitive engines.



\newpage

\bibliographystyle{IEEEtran}
\bibliography{bibliography}

\begin{thebibliography}{10}
\providecommand{\url}[1]{#1}
\csname url@samestyle\endcsname
\providecommand{\newblock}{\relax}
\providecommand{\bibinfo}[2]{#2}
\providecommand{\BIBentrySTDinterwordspacing}{\spaceskip=0pt\relax}
\providecommand{\BIBentryALTinterwordstretchfactor}{4}
\providecommand{\BIBentryALTinterwordspacing}{\spaceskip=\fontdimen2\font plus
\BIBentryALTinterwordstretchfactor\fontdimen3\font minus
  \fontdimen4\font\relax}
\providecommand{\BIBforeignlanguage}[2]{{%
\expandafter\ifx\csname l@#1\endcsname\relax
\typeout{** WARNING: IEEEtran.bst: No hyphenation pattern has been}%
\typeout{** loaded for the language `#1'. Using the pattern for}%
\typeout{** the default language instead.}%
\else
\language=\csname l@#1\endcsname
\fi
#2}}
\providecommand{\BIBdecl}{\relax}
\BIBdecl

\bibitem{mayoral2019industrial}
V.~Mayoral-Vilches, L.~U.~S. Juan, U.~A. Carbajo, R.~Campo, X.~S.
  de~C{\'a}mara, O.~Urzelai, N.~Garc{\'\i}a, and E.~Gil-Uriarte, ``Industrial
  robot ransomware: Akerbeltz,'' \emph{arXiv preprint arXiv:1912.07714}, 2019.

\bibitem{abadi2016tensorflow}
M.~Abadi, ``Tensorflow: learning functions at scale,'' in \emph{Proceedings of
  the 21st ACM SIGPLAN International Conference on Functional Programming},
  2016, pp. 1--1.

\bibitem{lockheed2011}
\BIBentryALTinterwordspacing
E.~Hutchins, M.~Cloppert, and R.~Amin, ``Intelligence-driven computer network
  defense informed by analysis of adversary campaigns and intrusion kill
  chains,'' \emph{Lockheed Martin Corporation White Paper}, 2011. [Online].
  Available:
  \url{https://www.lockheedmartin.com/content/dam/lockheed-martin/rms/documents/cyber/LM-White-Paper-Intel-Driven-Defense.pdf}
\BIBentrySTDinterwordspacing

\bibitem{moore2003metasploit}
H.~D. Moore, ``Metasploit framework,'' 2003, available at:
  \url{https://www.metasploit.com}.

\bibitem{ghanem2022hierarchical}
M.~C. Ghanem, T.~M. Chen, and E.~G. Nepomuceno, ``Hierarchical reinforcement
  learning for efficient and effective automated penetration testing of large
  networks,'' \emph{Journal of Intelligent Information Systems}, pp. 1--23,
  2022.

\bibitem{zennaro2020modeling}
F.~M. Zennaro and L.~Erdodi, ``Modeling penetration testing with reinforcement
  learning using capture-the-flag challenges: trade-offs between model-free
  learning and a priori knowledge,'' \emph{arXiv preprint arXiv:2005.12632},
  2020.

\bibitem{ou2005mulval}
X.~Ou, S.~Govindavajhala, A.~W. Appel \emph{et~al.}, ``Mulval: A logic-based
  network security analyzer.'' in \emph{USENIX security symposium},
  vol.~8.\hskip 1em plus 0.5em minus 0.4em\relax Baltimore, MD, 2005, pp.
  113--128.

\bibitem{ng2013pegasus}
A.~Y. Ng and M.~I. Jordan, ``Pegasus: A policy search method for large mdps and
  pomdps,'' \emph{arXiv preprint arXiv:1301.3878}, 2013.

\bibitem{applebaum2016intelligent}
A.~Applebaum, D.~Miller, B.~Strom, C.~Korban, and R.~Wolf, ``Intelligent,
  automated red team emulation,'' in \emph{Proceedings of the 32nd Annual
  Conference on Computer Security Applications}.\hskip 1em plus 0.5em minus
  0.4em\relax ACM, 2016, pp. 363--373.

\bibitem{zhou2019nig}
T.-y. Zhou, Y.-c. Zang, J.-h. Zhu, and Q.-x. Wang, ``Nig-ap: a new method for
  automated penetration testing,'' \emph{Frontiers of Information Technology \&
  Electronic Engineering}, vol.~20, no.~9, pp. 1277--1288, 2019.

\bibitem{ghanem2019reinforcement}
M.~C. Ghanem and T.~M. Chen, ``Reinforcement learning for efficient network
  penetration testing,'' \emph{Information}, vol.~11, no.~1, p.~6, 2019.

\bibitem{hu2020automated}
Z.~Hu, R.~Beuran, and Y.~Tan, ``Automated penetration testing using deep
  reinforcement learning,'' in \emph{2020 IEEE European Symposium on Security
  and Privacy Workshops (EuroS\&PW)}.\hskip 1em plus 0.5em minus 0.4em\relax
  IEEE, 2020, pp. 2--10.

\bibitem{maeda2021automating}
R.~Maeda and M.~Mimura, ``Automating post-exploitation with deep reinforcement
  learning,'' \emph{Computers \& Security}, vol. 100, p. 102108, 2021.

\bibitem{chen2023gail}
J.~Chen, S.~Hu, H.~Zheng, C.~Xing, and G.~Zhang, ``Gail-pt: An intelligent
  penetration testing framework with generative adversarial imitation
  learning,'' \emph{Computers \& Security}, vol. 126, p. 103055, 2023.

\bibitem{2018arXiv180606681A}
L.~{Alzola Kirschgens}, I.~{Zamalloa Ugarte}, E.~{Gil Uriarte}, A.~{Mu{\~n}iz
  Rosas}, and V.~{Mayoral Vilches}, ``{Robot hazards: from safety to
  security},'' \emph{ArXiv e-prints}, Jun. 2018.

\bibitem{vilches2019introducing}
V.~M. Vilches, L.~U.~S. Juan, B.~Dieber, U.~A. Carbajo, and E.~Gil-Uriarte,
  ``Introducing the robot vulnerability database (rvd),'' \emph{arXiv preprint
  arXiv:1912.11299}, 2019.

\bibitem{applebaum2017analysis}
A.~Applebaum, D.~Miller, B.~Strom, H.~Foster, and C.~Thomas, ``Analysis of
  automated adversary emulation techniques,'' in \emph{Proceedings of the
  Summer Simulation Multi-Conference}.\hskip 1em plus 0.5em minus 0.4em\relax
  Society for Computer Simulation International, 2017, p.~16.

\bibitem{ghanem2018reinforcement}
M.~C. Ghanem and T.~M. Chen, ``Reinforcement learning for intelligent
  penetration testing,'' in \emph{2018 Second World Conference on Smart Trends
  in Systems, Security and Sustainability (WorldS4)}.\hskip 1em plus 0.5em
  minus 0.4em\relax IEEE, 2018, pp. 185--192.

\bibitem{niculaelearning2018}
S.~Niculae, ``Applying reinforcement learning and genetic algorithms in
  game-theoretic cyber-security,'' p.~58, 2018.

\bibitem{schwartz2019autonomous}
J.~Schwartz and H.~Kurniawati, ``Autonomous penetration testing using
  reinforcement learning,'' \emph{arXiv preprint arXiv:1905.05965}, 2019.

\bibitem{mayoral2020alurity}
V.~Mayoral-Vilches, I.~Abad-Fern{\'a}ndez, M.~Pinzger, S.~Rass, B.~Dieber,
  A.~Cunha, F.~J. Rodr{\'\i}guez-Lera, G.~Lacava, A.~Marotta, F.~Martinelli
  \emph{et~al.}, ``alurity, a toolbox for robot cybersecurity,'' \emph{arXiv
  preprint arXiv:2010.07759}, 2020.

\end{thebibliography}

\FloatBarrier
\newpage

\appendices

\section{Code listings}

\definecolor{codegreen}{rgb}{0.3,0.6,0.3}
\definecolor{codegray}{rgb}{0.5,0.5,0.5}
\definecolor{codepurple}{rgb}{0.5,0,0.33}
\definecolor{backcolour}{rgb}{0.95,0.95,0.92}

\lstdefinestyle{mystyle}{
    backgroundcolor=\color{backcolour},
    commentstyle=\color{codegreen},
    keywordstyle=\color{blue},
    numberstyle=\tiny\color{codegray},
    stringstyle=\color{codepurple},
    basicstyle=\ttfamily\footnotesize,
    breakatwhitespace=false,
    breaklines=true,
    captionpos=b,
    keepspaces=true,
    numbers=left,
    numbersep=5pt,
    showspaces=false,
    showstringspaces=false,
    showtabs=false,
    tabsize=2
}
\lstset{style=mystyle,caption={Simplified listing of one of \exploitflow's \texttt{State} abstractions. Complete source code available at \faGithub~  \url{https://github.com/vmayoral/ExploitFlow/blob/main/exploitflow/state.py}}}
\begin{lstlisting}[language=Python, label=lst:state]
class State_v2(State):
    def __init__(self, *args):
        self.states = {}

        # initialize all states as empty
        for ip in TARGET_IP_ADDRESSES:
            self.add_new(ip)

    def merge(self, newstate, target="127.0.0.1") -> None:
        """
        Merges the current object with a new State

        Supports both State_v1 and State_v2.
        """
        if type(newstate) == State_v1:
            self.states[target] = newstate  # whether it exists or not            
        elif type(newstate) == State_v2:
            aux_state = self + newstate  # NOTE: overwrites self, with newstate
            self.states = aux_state.states
        else:
            raise TypeError("Unknown state type")

    # (...) various methods omitted
    
    def one_hot_encode(self):
        # one-hot encode all State_v1 objects in 'states'
        states_encoded = [state.one_hot_encode() for state in self.states.values()]
        flattened_states_encoded = [item for sublist in states_encoded for item in sublist]
        # return states_encoded
        return flattened_states_encoded 
\end{lstlisting}

\newpage

\lstdefinestyle{mystyle}{
    backgroundcolor=\color{backcolour},
    commentstyle=\color{codegreen},
    keywordstyle=\color{blue},
    numberstyle=\tiny\color{codegray},
    stringstyle=\color{codepurple},
    basicstyle=\ttfamily\footnotesize,
    breakatwhitespace=false,
    breaklines=true,
    captionpos=b,
    keepspaces=true,
    numbers=left,
    numbersep=5pt,
    showspaces=false,
    showstringspaces=false,
    showtabs=false,
    tabsize=2
}
\lstset{style=mystyle,caption={Q-Learning class implementing a common table-based Q-Learning algorithm. Simplified implementation removes comments and non-crucial methods. Complete source code available including comments and documentation at \faGithub~  \url{https://github.com/vmayoral/ExploitFlow/blob/main/exploitflow/models.py}}}
\begin{lstlisting}[language=Python, label=lst:qlearning]
class QLearn:
    """Q-Learning class. Implements the Q-Learning algorithm."""

    def __init__(self,
                 actions,
                 epsilon=0.1,
                 alpha=0.2,
                 gamma=0.9):
        self.q = {}
        self.epsilon = epsilon
        self.alpha = alpha
        self.gamma = gamma
        self.actions = actions

    def learnQ(self, state, action, reward, value, debug=False):
        """Updates the Q-value for a state-action pair.

        The core Q-Learning update rule.
            Q(s, a) += alpha * (reward(s,a) + max(Q(s')) - Q(s,a))
        """
        oldv = self.q.get((state, action), None)
        if oldv is None:
            self.q[(state, action)] = reward
        else:
            self.q[(state, action)] = oldv + self.alpha * (value - oldv)

    def chooseAction(self, state, return_q=False):
        """An alternative approach for action selection."""
        # Compute the Q values for each action given the current state
        q = [self.getQ(state, a) for a in self.actions]
        maxQ = max(q)
        if random.random() < self.epsilon:
            minQ = min(q)
            mag = max(abs(minQ), abs(maxQ))  # Determine the magnitude 
                                             # range based on minQ and maxQ

            q = [q[i] + random.random() * mag - .5 * mag for i in range(len(self.actions))]
            maxQ = max(q)
        count = q.count(maxQ)
        if count > 1:
            best = [i for i in range(len(self.actions)) if q[i] == maxQ]
            i = random.choice(best)
        else:
            i = q.index(maxQ)
        action = self.actions[i]
        if return_q:
            return action, q
        return action

    def learn(self, state1, action1, reward, state2, debug=False):
        """Get the maximum Q-Value for the next state."""
        maxqnew = max([self.getQ(state2, a) for a in self.actions])
        self.learnQ(state1, action1, reward, reward + self.gamma * maxqnew, debug=debug)
\end{lstlisting}

\newpage

\lstdefinestyle{mystyle}{
    backgroundcolor=\color{backcolour},
    commentstyle=\color{codegreen},
    keywordstyle=\color{blue},
    numberstyle=\tiny\color{codegray},
    stringstyle=\color{codepurple},
    basicstyle=\ttfamily\footnotesize,
    breakatwhitespace=false,
    breaklines=true,
    captionpos=b,
    keepspaces=true,
    numbers=left,
    numbersep=5pt,
    showspaces=false,
    showstringspaces=false,
    showtabs=false,
    tabsize=2
}
\lstset{style=mystyle,caption={Human-expert penetration tester using \exploitflow programatically programming an exploitation route for the best-case scenario. Complete experiment available at \faGithub~  \url{https://github.com/vmayoral/ExploitFlow/blob/main/examples/9_exploitation_ur_human_expert.py}}}
\begin{lstlisting}[language=Python, label=lst:human]
import exploitflow as ef
from exploitflow.state import State_v2
State_default = State_v2

flow = ef.Flow()
init = ef.Init()
recon = ef.Targets()
versions = ef.Versions(ports=ef.state.TARGET_PORTS_COMPLETE)
state = flow.run(init * recon * versions, target="192.168.2.10")

for s in state.states.keys():
    if any((port_state.port == 22 and port_state.open) for port_state in state.states[s].ports):
        expl = ef.adapter_msf_initializer.get_name("auxiliary", "scanner/ssh/ssh_login")
        msf_options = {
            "RHOSTS": s,
            "USERNAME": "root",
            "PASSWORD": "easybot"
        }        
        expl.set_options(msf_options)
        if not expl.missing():
            state = flow.run(state * expl, target=s, debug=False)
\end{lstlisting}

\newpage

\lstdefinestyle{mystyle}{
    backgroundcolor=\color{backcolour},
    commentstyle=\color{codegreen},
    keywordstyle=\color{blue},
    numberstyle=\tiny\color{codegray},
    stringstyle=\color{codepurple},
    basicstyle=\ttfamily\footnotesize,
    breakatwhitespace=false,
    breaklines=true,
    captionpos=b,
    keepspaces=true,
    numbers=left,
    numbersep=5pt,
    showspaces=false,
    showstringspaces=false,
    showtabs=false,
    tabsize=2
}
\lstset{style=mystyle,caption={Code snipped showing the training routine of an autonomous agent powered by Q-Learning. Complete experiment available at \faGithub~  \url{https://github.com/vmayoral/ExploitFlow/blob/main/examples/11_exploitation_ur_qlearning_instances.py}}}
\begin{lstlisting}[language=Python, label=lst:agent]
import exploitflow as ef
from exploitflow.state import State_v4
State_default = State_v4

flow = ef.Flow()
flow.set_learning_model(ef.QLearn(actions=exploits_encoded, alpha=0.1, gamma=0.9, epsilon=0.1))

rollouts = 1000
episode = 10
age = 1
debug = False
last_10_actions = []
while age <= rollouts:
    if flow.last_state():
        flow._graph.learning_model.learn(
            tuple(flow.last_state().one_hot_encode()), 
            flow.last_action().name, 
            flow.last_reward(), 
            tuple(flow.state().one_hot_encode()),
            debug=False)
        
    action = flow._graph.learning_model.chooseAction(tuple(flow.state().one_hot_encode()))

    flow.run(flow.state() * action_expl, debug=debug)
    
    if age % episode == 0:        
        # reset the flow
        flow.reset()

    # next rollout
    age += 1
\end{lstlisting}

\newpage

\lstdefinestyle{mystyle}{
    backgroundcolor=\color{backcolour},
    commentstyle=\color{codegreen},
    keywordstyle=\color{blue},
    numberstyle=\tiny\color{codegray},
    stringstyle=\color{codepurple},
    basicstyle=\ttfamily\footnotesize,
    breakatwhitespace=false,
    breaklines=true,
    captionpos=b,
    keepspaces=true,
    numbers=left,
    numbersep=5pt,
    showspaces=false,
    showstringspaces=false,
    showtabs=false,
    tabsize=2
}
\lstset{style=mystyle,caption={Code snipped showing the brute-forcing effort of the scenario, trying all possible permutations of the exploits (a reduced set for computational reasons). Complete experiment available at \faGithub~  \url{https://github.com/vmayoral/ExploitFlow/blob/main/examples/13_exploitation_ur_bruteforce.py}}}
\begin{lstlisting}[language=Python, label=lst:brute]
import exploitflow as ef
from exploitflow.state import State_v4
State_default = State_v4

flow = ef.Flow()
exploits = [ef.idle, ef.metasploit, ef.versions, ef.targets]

# Get all permutations of the list
permutations = list(itertools.permutations(exploits))
state = flow.run(ef.init)
for perm in permutations:
    for expl in perm:
        state = flow.run(flow.state() * expl)
\end{lstlisting}

\FloatBarrier
\newpage

\section{Large Language Models (LLMs) experiments to generate exploits in the desired format}

\lstdefinestyle{mystyle}{
    backgroundcolor=\color{backcolour},
    commentstyle=\color{codegreen},
    keywordstyle=\color{blue},
    numberstyle=\tiny\color{codegray},
    stringstyle=\color{codepurple},
    basicstyle=\ttfamily\footnotesize,
    breakatwhitespace=false,
    breaklines=true,
    captionpos=b,
    keepspaces=true,
    numbers=left,
    numbersep=5pt,
    showspaces=false,
    showstringspaces=false,
    showtabs=false,
    tabsize=2
}
\lstset{style=mystyle,caption={Code snipped showing the results of using generative Large Language Models to economically address the scalability problem of generating usable datasets and exploits for the training. The snippet shows how ChatGPT (GPT-4) is able to easily provide exploits for direct use.}}
\begin{lstlisting}[language=Python, label=lst:llm]

# ChatGPT (GPT-4 model) PROMPT:
##############################
#
# Select 10 exploits from the metasploit framework (MSF) and for each, produce an instances of it fillling up the corresponding options required for each exploit while adhering to the following format:
#
# metasploit_1 = ef.adapter_msf_initializer.get_name("auxiliary", "scanner/ssh/ssh_login")
# metasploit_1.name = "scanner/ssh/ssh_login @ 192.168.2.5"
# msf_options = {
#     "RHOSTS": "192.168.2.5",
#     "USERNAME": "root",
#     "PASSWORD": "easybot"
# }
# metasploit_1.set_options(msf_options)
# metasploit_1.target = "192.168.2.5"
# metasploit_1.reward = -100   # first time successful, will do "*(-1)" to get "+100   
#
# Provide the code necessary to instantiate the 10 exploits. For each exploit, at the beginning, provide a small comment of documentation describing what each exploit is expected to do.
##############################

# 1. This auxiliary module will attempt to login to a SSH service with a specific username and password.
metasploit_1 = ef.adapter_msf_initializer.get_name("auxiliary", "scanner/ssh/ssh_login")
metasploit_1.name = "scanner/ssh/ssh_login @ 192.168.2.5"
msf_options_1 = {
    "RHOSTS": "192.168.2.5",
    "USERNAME": "root",
    "PASSWORD": "easybot"
}
metasploit_1.set_options(msf_options_1)
metasploit_1.target = "192.168.2.5"
metasploit_1.reward = -100

# 2. This module exploits a buffer overflow in Apache's mod_isapi for versions 2.2.14 and earlier.
metasploit_2 = ef.adapter_msf_initializer.get_name("exploit", "windows/http/apache_mod_isapi")
metasploit_2.name = "windows/http/apache_mod_isapi @ 192.168.2.6"
msf_options_2 = {
    "RHOSTS": "192.168.2.6",
    "TARGETURI": "/somepath/"
}
metasploit_2.set_options(msf_options_2)
metasploit_2.target = "192.168.2.6"
metasploit_2.reward = -100

# 3. This auxiliary module will scan for vulnerable open FTP servers.
metasploit_3 = ef.adapter_msf_initializer.get_name("auxiliary", "scanner/ftp/ftp_version")
metasploit_3.name = "scanner/ftp/ftp_version @ 192.168.2.7"
msf_options_3 = {
    "RHOSTS": "192.168.2.7"
}
metasploit_3.set_options(msf_options_3)
metasploit_3.target = "192.168.2.7"
metasploit_3.reward = -100

# 4. This module exploits a directory traversal vulnerability in PHPMyAdmin.
metasploit_4 = ef.adapter_msf_initializer.get_name("exploit", "multi/http/phpmyadmin_3522_backdoor")
metasploit_4.name = "multi/http/phpmyadmin_3522_backdoor @ 192.168.2.8"
msf_options_4 = {
    "RHOSTS": "192.168.2.8",
    "TARGETURI": "/phpmyadmin/"
}
metasploit_4.set_options(msf_options_4)
metasploit_4.target = "192.168.2.8"
metasploit_4.reward = -100

# 5. This module exploits a buffer overflow in the Microsoft IIS FTP service.
metasploit_5 = ef.adapter_msf_initializer.get_name("exploit", "windows/ftp/ms09_053_ftpd_nlst")
metasploit_5.name = "windows/ftp/ms09_053_ftpd_nlst @ 192.168.2.9"
msf_options_5 = {
    "RHOSTS": "192.168.2.9"
}
metasploit_5.set_options(msf_options_5)
metasploit_5.target = "192.168.2.9"
metasploit_5.reward = -100

# 6. This auxiliary module will scan the network for Oracle servers.
metasploit_6 = ef.adapter_msf_initializer.get_name("auxiliary", "scanner/oracle/oracle_login")
metasploit_6.name = "scanner/oracle/oracle_login @ 192.168.2.10"
msf_options_6 = {
    "RHOSTS": "192.168.2.10"
}
metasploit_6.set_options(msf_options_6)
metasploit_6.target = "192.168.2.10"
metasploit_6.reward = -100

# 7. This module exploits a vulnerability in Microsoft's Server service.
metasploit_7 = ef.adapter_msf_initializer.get_name("exploit", "windows/smb/ms08_067_netapi")
metasploit_7.name = "windows/smb/ms08_067_netapi @ 192.168.2.11"
msf_options_7 = {
    "RHOSTS": "192.168.2.11"
}
metasploit_7.set_options(msf_options_7)
metasploit_7.target = "192.168.2.11"
metasploit_7.reward = -100

# 8. This module exploits a vulnerability in Windows' handling of VCard files.
metasploit_8 = ef.adapter_msf_initializer.get_name("exploit", "windows/fileformat/ms10_026_dbldecode")
metasploit_8.name = "windows/fileformat/ms10_026_dbldecode @ 192.168.2.12"
msf_options_8 = {
    "RHOSTS": "192.168.2.12"
}
metasploit_8.set_options(msf_options_8)
metasploit_8.target = "192.168.2.12"
metasploit_8.reward = -100

# 9. This module exploits a buffer overflow in the Samba service.
metasploit_9 = ef.adapter_msf_initializer.get_name("exploit", "unix/misc/distcc_exec")
metasploit_9.name = "unix/misc/distcc_exec @ 192.168.2.13"
msf_options_9 = {
    "RHOSTS": "192.168.2.13"
}
metasploit_9.set_options(msf_options_9)
metasploit_9.target = "192.168.2.13"
metasploit_9.reward = -100

# 10. This module exploits a vulnerability in the Windows handling of HLP files.
metasploit_10 = ef.adapter_msf_initializer.get_name("exploit", "windows/fileformat/ms10_042_helpctr_xss_cmd_exec")
metasploit_10.name = "windows/fileformat/ms10_042_helpctr_xss_cmd_exec @ 192.168.2.14"
msf_options_10 = {
    "RHOSTS": "192.168.2.14"
}
metasploit_10.set_options(msf_options_10)
metasploit_10.target = "192.168.2.14"
metasploit_10.reward = -100


\end{lstlisting}

\end{document}